# American Sign Language to Text Translation using Transformer and Seq2Seq with LSTM

Gregorius Guntur Sunardi Putra
*Department of Informatics*
*Institut Teknologi Sepuluh Nopember*
Surabaya, Indonesia
6025231023@student.its.ac.id

Adifa Widyadhani Chanda D'Layla
*Department of Informatics*
*Institut Teknologi Sepuluh Nopember*
Surabaya, Indonesia
5025201013@student.its.ac.id

Dimas Wahono
*Department of Informatics*
*Institut Teknologi Sepuluh Nopember*
Surabaya, Indonesia
6025231048@student.its.ac.id

Riyanarto Sarno
*Department of Informatics*
*Institut Teknologi Sepuluh Nopember*
Surabaya, Indonesia
riyanarto@its.ac.id

Agus Tri Haryono
*Department of Informatics*
*Institut Teknologi Sepuluh Nopember*
Surabaya, Indonesia
7025231020@student.its.ac.id

*Abstract*—Sign language translation is one of the important issues in communication between deaf and hearing people, as it expresses words through hand, body, and mouth movements. American Sign Language is one of the sign languages used, one of which is the alphabetic sign. The development of neural machine translation technology is moving towards sign language translation. Transformer became the state-of-the-art in natural language processing. This study compares the Transformer with the Sequence-to-Sequence (Seq2Seq) model in translating sign language to text. In addition, an experiment was conducted by adding Residual Long Short-Term Memory (ResidualLSTM) in the Transformer. The addition of ResidualLSTM to the Transformer reduces the performance of the Transformer model by 23.37% based on the BLEU Score value. In comparison, the Transformer itself increases the BLEU Score value by 28.14 compared to the Seq2Seq model.

## I. INTRODUCTION

Sign language is a method of communication of the deaf community with hearing friends who communicate in spoken language. However, the many accents or characteristics of sign language make communication require an interpreter. Sign language utilizes hand, body, and lip movements to indicate words, alphabets, and numbers [1]. Sign language interpretation is needed not only for communication between hearing and deaf but also among deaf communities. That is due to the many differences in sign language in each region despite adopting the same language. There are currently over 135 different sign languages around the world, and this does not include regional sign languages [2].

American Sign Language (ASL) is the dominant sign language in the United States. More than 20 countries use ASL for hearing-impaired communication [3]. ASL has many different dialects, depending on pronunciation, sounds, and slang. ASL became a reference in creating the Indonesian Sign Language System (SIBI) by modifying ASL grammar into Indonesian. Fingerspelling is a common sign language used for signing proper nouns and when the sign of an object is unknown. Fingerspelling translates a word into a common standard character sign. Fingerspelling has 26 gesture sign sets for alphabet signs and ten gesture sets for number signs. Its standardized nature makes it easy to create automatic sign language translation that facilitates deaf-to-deaf and deaf-to-hearing communication.

Long Short-Term Memory (LSTM) [4], is a variation of Recurrent Neural Networks (RNN) that has been successfully applied to various Natural Language Processing (NLP) tasks. LSTM has the ability to capture long-range dependencies with sequences making it suitable for solving sequence labeling problems [5], [6], [7], such as Named Entity Recognition (NER), Part-of-Speech tagging (POS-Tag) and Neural Machine Translation (NMT). Study [8] combining Bidirectional LSTM (BiLSTM) model with Conditional Random Field (CRF) to solve NER and POS-Tag Task. BiLSTM encodes word or character vectors into features, while CRF is a classifier to predict the target labels. Previous research has explored enhancing the feature extractor by integrating serves attention mechanisms [9] with existing models or introducing alternative extractors such as Convolutional Neural Networks (CNN) [10], Transformer [11], [12], and Graph Neural Networks [13]. Additionally, some models have aimed to incorporate additional semantic information, such as pre-trained language models, segmentation information, and contextual cues.

Increasing its depth is one viable method to enhance neural network performance, as demonstrated effectively in Convolutional Neural Networks (CNNs). For instance, ResNets and DenseNets have achieved impressive results in image recognition tasks by constructing deep networks using identity skip connections [14], [15]. These connections help address the vanishing gradient problem in deep networks and allow for the reuse of features from different layers, thereby improving overall performance. Consequently, the residual structure has become a crucial element widely incorporated into many neural network models. Several stacked Residual LSTM models have been proposed for Natural Language Processing (NLP) tasks, indicating that LSTM networks can also benefit from residual connections [16].

This paper proposes translating American Sign Language to English using the Transformer. Residual LSTM will be combined into the Transformer especially in Landmarks Embedding. The Transformer model and Residual LSTM-Transformer will be compared with the Seq2Seq model. The forthcoming segments of this study are divided into four sections: Section II, which delves into existing literature; Section III, which elaborates on the proposed methodology. Section IV encompasses the attained results, and Section V wraps up this article.

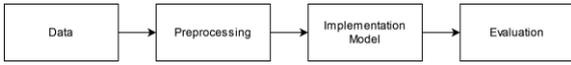

Fig. 1. Flowchart system machine translation

## II. RELATED WORKS

In the field of image recognition, VLAD [17] represents a method that encodes residual vectors relative to a dictionary, while Fisher Vector [18] can be seen as a probabilistic adaptation of VLAD[17]. Both serve as robust shallow representations for image retrieval and classification tasks. Studies have demonstrated that encoding residual vectors proves more effective than encoding original vectors for vector quantization. Solving Partial Differential Equations (PDEs) in low-level vision and computer graphics often involves methods like Multigrid, which decomposes the system into subproblems at multiple scales, each addressing the residual solution between coarser and finer scales. An alternative approach is hierarchical basis preconditioning, which utilizes variables representing residual vectors[19] between scales. Research indicates that these methods converge much faster than standard solvers that do not consider the residual nature of solutions, underscoring the importance of effective reformulation or preconditioning in optimization tasks.

A previous study analyzed text-based data to extract opinions from teaching staff members [20]. They achieved this by evaluating the performance of the best teacher among the entire staff using an attention-based LSTM model. Another author proposed a model combining CNN and bidirectional gated recurrent units to detect semantic nuances within sentences [21]. Another approach introduced a lexicon-based enhanced LSTM model, utilizing pre-trained lexicon information for classification purposes [22]. Another model utilized pattern recognition algorithms to identify sarcasm in Twitter data, distinguishing four distinct types of sarcasm based on specific features [23].

In [12] addresses large-scale image search, focusing on accuracy, efficiency, and memory usage, demonstrating that the Fisher kernel outperforms the bag-of-visual-words approach. By optimizing dimensionality reduction and indexing, it enables compact and accurate image representation, allowing the search of 100 million images to be completed in approximately 250 ms on a single processor core. In [16] introduces a novel low-complexity model called CNN-BiGRU, which combines convolutional neural networks and bidirectional gated recurrent units to improve sentence semantic classification by focusing on both keywords and underlying semantics. Comprehensive experiments on seven benchmarks demonstrate that this model outperforms state-of-the-art methods by effectively integrating contextual representations and semantic distribution, learning a compact code for sentence sentiment classification with limited hyper-parameters.

In the study [24] introduced five BERT-based pre-trained Transformer models aimed at identifying suicidal intentions. These pre-trained models, namely RoBERTa, ALBERT, BERT, DistilBERT, and DistilRoBERTa, were applied to a dataset of tweets containing 9119 entries categorized into intention and no intention. The tweets underwent preprocessing steps, including fixing broken Unicode, converting capital letters to lowercase, and mapping contractions. Subsequently, the data was split into training, validation, and testing sets using stratified random sampling. The five Transformer models were then employed to detect suicide intentions within the datasets. Among these models, RoBERTa demonstrated superior performance across all stages, achieving scores of 99.23%, 96.35%, and 95.39% for training, validation, and testing, respectively.

This research aims to introduce a novel sign language translation scheme to enhance accuracy and comprehension capabilities. The proposed method is inspired by the insights from previous works [12], [16] discussed in this section and utilizes the same dataset [25] as used in prior studies.

## III. METHODOLOGY

This section introduces the dataset, model sign language, experimental setting, and evaluation model. The flowchart in Figure 1 illustrates the system's process of translating and processing sign language.

### A. Dataset

The dataset uses the American Sign Language Recognition Dataset [25]. This dataset consists of face, right-hand, and left-hand landmarks. There are 52,958 data divided into 53 parquet data. The data has landmarks taken from sign language videos using media pipes as sources. In addition, 503 unique phrases with a range of 16 - 43 characters are used as targets. The sample phrase data is shown in Table I, and the distribution of char length is shown in Figure 2. The sample left-hand landmark is shown in Figure 3.

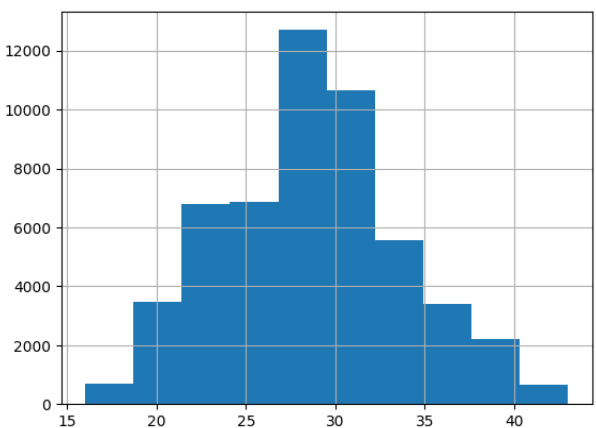

Fig. 2. Distribution Length Character of Dataset

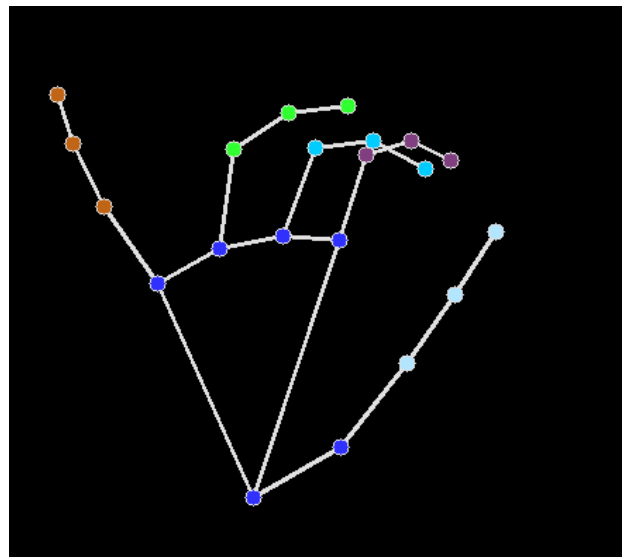

Fig. 3. Left-hand landmark

TABLE I. SAMPLE OF THE TEXT TARGET

| Phrase | Char |
|---|---|
| i agree with you | 16 |
| i hate baking pies | 18 |
| take a coffee break | 19 |
| employee recruitment takes a lot of effort | 42 |
| if you come home late the doors are locked | 42 |
| a steep learning curve in riding a unicycle | 43 |

*B. Preprocessing Data*

Data from sign language and target text will be preprocessed by embedding to be input into Transformer. The embedding process converts data into vectors. In building machine translation, embedding is done to convert words into vectors. In this study will convert frames and phrase tokens into vectors. Landmark embedding utilizes three convolutional layers to get feature from Landmarks MediaPipe results. Character level tokens will be used as embedding tokens. Landmarks embedding will be used as the source and becomes the input for Transformer Encoder, while Token Embedding becomes the input for Transformer Decoder. The data results are then divided into training, validation, and testing. Training data uses 80% of the total data, validation data uses 20%, while testing data for evaluation uses 2,000 data from validation data.

*C. Seq2Seq Translation Model*

Sequence to sequence requires encoder and decoder networks to process language translation. The process begins with encoding source sequences to a fixed-sized vector, which is then decoded to generate the target sequences [26].

The Seq2Seq architecture uses an encoder and a decoder, utilizing LSTM and added dropout with a probability value of 0.5 to reduce complexity. The encoder uses an LSTM with a unit of 10. The decoder uses an LSTM with a unit of 62, with the encoder's hidden states as initial states, and generates the sequence of translation output. The overall architecture of the model is shown in Figure 4.

*D. Transformer*

One of the most important concepts of the Transformer model is multi-head attention [12]. In multi-head attention, there are four layers: input, dot-product attention, concate, and output. Input consists of 3 dense layers that receive queries, keys, and values. The output of the dot-product is transformed into linear projection, which produces output as a head. In machine translation, a causal attention mask is added and used

```
Algorithm 1 Seq2Seq Model
Encoder:
    Dropout Layer: Dropout(p=0.5, inplace=False)
    Embedding Layer: LandmarkEmbedding
        Conv1: Conv1d(64, 64, kernel_size=(11,), stride=(2,), padding=(5,))
        Conv2: Conv1d(64, 64, kernel_size=(11,), stride=(2,), padding=(5,))
        Conv3: Conv1d(64, 64, kernel_size=(11,), stride=(2,), padding=(5,))
        Positional Embedding: Embedding(100, 64)
        ReLU: ReLU()
    RNN: LSTM(10, 1024, num_layers=2, dropout=0.5)

Decoder:
    Dropout Layer: Dropout(p=0.5, inplace=False)
    Embedding Layer: Embedding(62, 62)
    RNN: LSTM(62, 1024, num_layers=2, dropout=0.5)
    Fully Connected: Linear(in_features=1024, out_features=62, bias=True)
```

Fig. 4. Pseudocode Seq2Seq

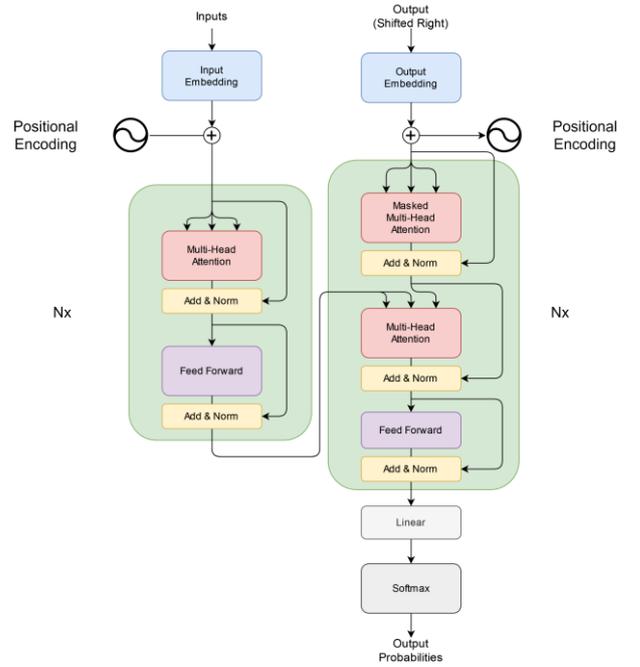

Fig. 5. Archiitecture Transformer

in the decoder Transformer to restrict the model from looking forward. It simply ensures that the prediction result of the next word depends on the previous words. The Transformer assigns different levels of importance to each input using a self-attention mechanism.

Fig. 5 shows the general architecture of the Transformer. This study uses an encoder consisting of multi-head attention and residual normalization. The Encoder Transformer in this study used five layers, while the decoder used one layer. The parameter on the Transformer model uses num_hid 100, head 4, feed-forward 40, frame max len 128, target max len 64, encoding layer 5, decoding layer 1, and class 64. Sixty-four classes are classes with letters, numbers, unique characters, and three unique tokens as start, end, and pad tokens. The loss function uses categorical cross entropy and Levantine Distance. The optimizer uses Adam with a value of 0.001.

*E. Residual LSTM - Transformer*

Residual LSTM overcomes the challenge of gradient loss in deep LSTM by providing information through two paths. One path updates temporal data like a regular LSTM, while the other path acts as a shortcut, directly connecting the previous layer to the current layer [16]. This separation allows the LSTM cell to organize its internal flow without interfering

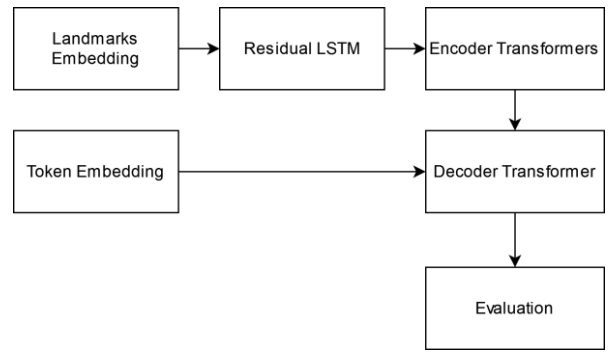

Fig. 6. Flowchart Model ResidualLSTM in Transformer

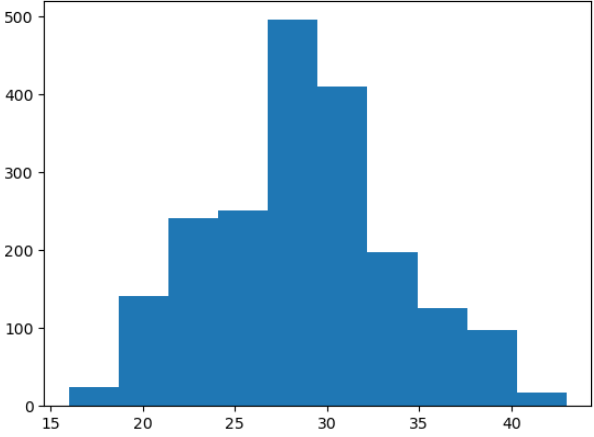

Fig. 7. Distribution of Test Data

with the shortcut path. Residual LSTM is efficient in handling gradients.

This study combines Residual LSTM and Transformer models to achieve better performance in sign language recognition. Our proposed model is shown in Figure 6 Residual LSTM is used as an additional embedding of landmark embedding. The Residual LSTM architecture consists of one LSTM and two Dense layers, and it uses ReLU as the activation function.

*F. Experimental setup*

The models were trained using the Kaggle Notebook with P100 GPU kernels 16GB VRAM. The Seq2Seq model was built using PyTorch frameworks, while the Transformer used Tensorflow frameworks. The Adam optimizer was used to train the weights, and an epoch value of 50 was set with an initial learning rate of 0.001. The batch size value was set to 64. Loss values are calculated using categorical cross entropy and Levenshtein Distance values.

*G. BLUE Score and CER Score*

The BLEU (Bilingual Evaluation Understudy) score assesses machine translation quality. BLEU compares the machine's attempt with a human-made translation, giving a score between 0 and 1. That metric counts the frequency of cooccurring word tokens in two sentences using Eq. 1. Penalty factor *(BP)* using *1* when the length of translation sentence ($l_t$) is more than the actual sentence ($l_a$). Otherwise, the *BP* value obtained from $e^{1-l_a/l_t}$. *Candidates* are the set of all sentences that need to be translated, $n-gram$ represent a string of $n$ consecutive words, and $count_{clip}(n-gram)$ is the minimum number of occurrences of $n-gram$. $P_n$ shows the similarity between translation and actual sentence, that value calculated from Eq. 2. Weight value ($\omega$) is used for how much BLEU is used and by default, BLEU-4 is 0.25 for ($\omega_1$ to $\omega_4$). In this study, $\omega$ value using $\omega_1 = 1$. A higher score means the machine translation is closer to what a human translator would produce.

$$BLEU = BP * \exp\left(\sum_{n=1}^{N} \omega_n \log P_n\right) \quad (1)$$

$$Pn = \frac{\sum_{C \in \{Candidates\}} \sum_{n-gram \in C} count_{clip}(n-gram)}{\sum_{C' \in \{Candidates\}} \sum_{n-gram' \in C} count(n-gram')} \quad (2)$$

Word Error Rate is a metric used to measure machine translation performance. It assesses how many letter errors there are in a sentence. Like calculating the BLUE score, the target and predicted will be tokenized word level. Next, the number of substitutions (*S*), deletions (*D*), insertions (*I*), and correct words (*W*) are calculated. Another one that can use WER for character level is called CER. WER can be computed using Eq. 3.

$$WER = \frac{S+D+I}{S+D+W} * 100\% \quad (3)$$

IV. RESULTS AND DISCUSSION

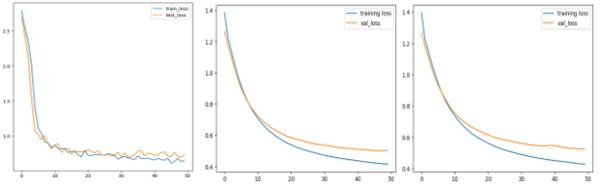

Fig. 8. Loss Progression over 50 epochs in Seq2Seq (left), Transformer with Residual-LSTM (mid), and Transformer (right).

This section focuses more on discussing the experiment results of the implementation work. The distribution of the test data used for model evaluation is shown in Figure 7. The test data represents 2,000 data from the validation data. The distribution of the data is similar to that of the test data.

TABLE II. COMPARISON OF DIFFERENT MODELS

| Model | BLEU | WER |
|---|---|---|
| Seq2Seq | 0.5687 | 74.56% |
| Residual LSTM Transformer | 0.6164 | 100.27% |
| **Transformer** | **0.8501** | **37.62%** |

The sign language translation to text uses deep learning models with three different models: LSTM-based Sequence-to-Sequence for encoder and decoder, Transformer, and Transformer with Residual LSTM. In the ASL data set, 80% of the data is used as training data, the remaining 20% is used as test data, and the validation data uses 50% of the test data. Figure 8 shows the progression of loss over 50 epochs for each translation model. The graph shows that after training the model for 50 epochs, the final training loss achieved was 0.6566 for the Seq2Seq model, 0.4540 for the Transformer model, and 0.5068 for the Transformer with Residual LSTM.

TABLE III. EXAMPLE OF TRANSLATION WITH HIGHEST BLUE SCORE

| No | Actual | Prediction | Char | BLEU |
|---|---|---|---|---|
| 1 | microscopes make small things look big | microscopes make small things look big | 45 | 1.0 |
| 2 | be persistent to win a strike | be persistent to win a strike | 37 | 1.0 |
| 3 | prescription drugs require a note | prescription drugs require a note | 35 | 1.0 |
| 4 | an excellent way to communicate | an excellent way to communicate | 31 | 1.0 |
| 5 | but the levee was dry | but the levee was dry | 29 | 1.0 |
| 6 | peek out the window | peek out the window | 21 | 1.0 |
| 7 | the biggest hamburger i have ever seen | the biggest hamburger i heve ever seen | 45 | 0.9736 |
| 8 | the living is easy | the living is sea | 21 | 0.9428 |
| 9 | they might find your comment offensive | the kis my find your comment offensive | 45 | 0.9210 |
| 10 | i took the rover from the shop | i took the rover from the shopent | 37 | 0.9090 |

Each sign language translation model is evaluated based on Table II WER and BLEU Score values. Based on the BLEU value in Table II, the Transformer model is better than the Seq2Seq model, with an increase of 28.14% for the baseline Transformer and 4.77% for the Transformer with Residual LSTM. The smaller the WER value, the better the model's performance in handling errors. Adding Residual

LSTM to the Transformer model resulted in many wrong words regarding substitutions, deletions, and insertions, which was 25.71% worse than the Seq2Seq model. When looking back at the evaluation value between BLEU and WER, there is a considerable difference in results, this is because BLEU calculates the overlap value between predicted and actual results, while WER considers word error in calculating overlap.

The model evaluation results show that the Transformer model is better than the other two. Some sentence results of the Transformer model translation with the best BLEU Score from the test data are shown in Table III, and the sentence results of the Transformer model translation with the lowest BLEU Score from the test data are shown in Table IV.

In testing, the Transformer model was found to have a satisfactory BLEU Score, especially when the sentences were 21-37 characters. Our observations show that data of a certain length shows semantic sentence errors, which needs attention to determine the model's ability to recognize the long-term context. In Table III, the 7th to 10th sentences show similarity between the predicted and actual sentences with one or two-word errors. Table IV shows overall translation errors where the model does not recognize the form of words and sentences. That is because the data with a specific length used in the training process is still limited, especially for data with character lengths below 21 and above 37. In terms of average BLEU Score, the low freq data have avg little different between high frequency data as show on Figure 9. In addition, the translation model uses the character level to overcome the out-of-vocabulary problem, so it cannot recognize the complete word.

TABLE IV. EXAMPLE OF TRANSLATION WITH LOWEST BLUE SCORE

| No | Actual | Prediction | Char | BLEU |
|---|---|---|---|---|
| 1 | the water was monitored daily | he is angany | 37 | 0.2021 |
| 2 | dormitory doors are locked at midnight | ned any ouang thogh | 45 | 0.2710 |
| 3 | this equation is too complicated | the sedratic arde | 37 | 0.3573 |
| 4 | fish are jumping | he inng othe an the oik | 21 | 0.3913 |
| 5 | sprawling subdivisions are bad | the aresperive is has before a efe a effort | 37 | 0.4285 |
| 6 | companies announce a merger | be doilst for a go fail | 29 | 0.4384 |
| 7 | dinosaurs have been extinct ages | trepangule be of extjume | 45 | 0.4801 |
| 8 | life is but a dream | love like th lthe liss bert | 21 | 0.4814 |
| 9 | safe to walk the streets in the evening | i fiets do be syour a hese | 45 | 0.4898 |
| 10 | i am allergic to bees and peanuts | what o kay o sthogsediates | 37 | 0.4995 |

## V. CONCLUSION

This paper compares the Transformer model with Seq2Seq to translate sign language into text. Sign language translation using Transformer has performed better than Seq2Seq based on BLEU and WER Score values. The Transformer model shows a 28.14% performance improvement over Seq2Seq. In addition, the Transformer model shows better translation quality. However, the translation results obtained are still unsatisfactory whereas the translation in the form of links is still not well translated. In the future, the minority number of char length problems is an emergency task, so need to handle low frequency char problem. In addition, Transformer can be

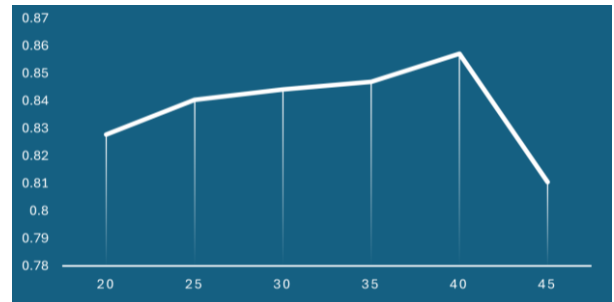
Fig. 9. Average BLEU-1 in Transformer Model

tested for unstructured sign languages at the word level and other sign languages such as BISINDO.

ACKNOWLEDGMENT

This research was funded by Institut Teknologi Sepuluh Nopember (ITS) under Penelitian Dana Departemen Program, Penelitian Keilmuan, Penelitian Flagship, Penelitian Kolaborasi Pusat and under the Project Scheme of the Publication Writing and Intellectual Property Rights (IPR) Incentive (Penulisan Publikasi dan Hak Kekayaan Intelektual/PPHKI) Program 2024.